\documentclass{bmvc2k}

\title{Bias-Awareness for Zero-Shot Learning\\ the Seen and Unseen}

\addauthor{William Thong}{w.e.thong@uva.nl}{1}
\addauthor{Cees G. M. Snoek}{cgmsnoek@uva.nl}{1}

\addinstitution{
 University of Amsterdam\\
 Science Park 904\\
 Amsterdam\\
 The Netherlands
}

\runninghead{Thong and Snoek}{Bias-Awareness for Generalized Zero-Shot Learning}

\def\eg{\emph{e.g}\bmvaOneDot}

\def\etal{\emph{et al}\bmvaOneDot}

\usepackage{amsmath}
\usepackage{arydshln}
\usepackage{amsfonts}       
\usepackage{nicefrac}       
\usepackage{microtype}      
\usepackage{booktabs}
\usepackage{multirow}
\usepackage{dirtytalk}
\usepackage{comment}
\usepackage{overpic}

\DeclareMathOperator{\unseen}{unseen}
\DeclareMathOperator{\seen}{seen}

\definecolor{Gray}{gray}{0.9}
\newcommand{\deemph}[1]{\textcolor{gray}{#1}}

\newcolumntype{x}[1]{>{\centering\arraybackslash}p{#1pt}}
\newlength\savewidth
\newcommand{\tablestyle}[2]{\setlength{\tabcolsep}{#1}\renewcommand{\arraystretch}{#2}\centering\small}

\makeatletter
\renewcommand\paragraph{\@startsection{paragraph}{4}{\z@}%
  {.5em \@plus1ex \@minus.5ex}%
  {-.5em}%
  {\normalfont\normalsize\bfseries\textcolor{bmv@captioncolor}}}
\makeatother

\begin{document}

\maketitle

\begin{abstract}
Generalized zero-shot learning recognizes inputs from both seen and unseen classes. Yet, existing methods tend to be biased towards the classes seen during training. In this paper, we strive to mitigate this bias. We propose a bias-aware learner to map inputs to a semantic embedding space for generalized zero-shot learning. During training, the model learns to regress to real-valued class prototypes in the embedding space with temperature scaling, while a margin-based bidirectional entropy term regularizes seen and unseen probabilities. Relying on a real-valued semantic embedding space provides a versatile approach, as the model can operate on different types of semantic information for both seen and unseen classes. Experiments are carried out on four benchmarks for generalized zero-shot learning and demonstrate the benefits of the proposed bias-aware classifier, both as a stand-alone method or in combination with generated features.
\end{abstract}

\section{Introduction}
\label{sec:intro}

Zero-shot recognition~\cite{lampert2014attribute,palatucci2009zero} considers if models trained on a given set of $\seen$ classes $\mathcal{S}$ can extrapolate to a distinct set of $\unseen$ classes $\mathcal{U}$. In generalized zero-shot learning~\cite{chao2016empirical,xian2018zero}, we also want to remember the $\seen$ classes and evaluate over the union of the two sets of classes $\mathcal{T}=\mathcal{S}\cup\mathcal{U}$.
Nevertheless, when evaluating existing models in the generalized scenario, the seminal work of Chao~\etal~\cite{chao2016empirical} highlights that predictions tend to be biased towards the $\seen$ classes observed during training. In this paper, we consider the challenge of mitigating this inherent bias present in classifiers by proposing a bias-aware model.

An effective remedy to remove the bias towards seen classes is to calibrate their predictions during inference. Chao~\etal~\cite{chao2016empirical} propose to reduce the scores for the seen classes, which in return improves the generalized zero-shot learning performance. Yet, the bias towards seen classes should also be tackled while training classifiers, and not only during the evaluation phase, to address the bias from the start.
Towards this goal, seen and unseen classes can be addressed separately during training. Liu~\etal~\cite{liu2018generalized} define two separate training objectives to calibrate the confidence of seen classes and the uncertainty of unseen classes.
Atzmon and Chechik~\cite{atzmon2018domain} break the classification into two separate experts, with one model for seen classes and another one for unseen classes. Their COSMO approach provides compelling results at the expense of a third additional expert to combine results. As generalized zero-shot learning considers both seen and unseen classes simultaneously, learners should benefit from mitigating the bias in both directions by considering both sets jointly rather than separately.

The main objective of this paper is to mitigate the bias towards seen classes by considering predictions of seen and unseen classes simultaneously during training. To achieve this, we propose a simple bias-aware learner that maps inputs to a semantic embedding space where class prototypes are formed by real-valued representations. We address the bias by introducing (\textit{i}) a calibration for the learner with temperature scaling, and (\textit{ii}) a margin-based bidirectional entropy term to regularize seen and unseen probabilities jointly.
We show that the bias towards seen classes is also dataset-dependent, and every dataset does not suffer to the same extent.
Finally, we illustrate the versatility of our approach. By relying on a real-valued embedding space, the model can handle different types of prototype representation for both seen and unseen classes, and operate either on real features, akin to compatibility functions, or leverage generated unseen features. Comparisons on four datasets for generalized zero-shot learning show the effectiveness of bias-awareness. All source code and setups are released\footnote{Source code is available at \href{https://github.com/twuilliam/bias-gzsl}{https://github.com/twuilliam/bias-gzsl}}.

\section{Related Work}\label{sec:rw}

\paragraph{Generalized zero-shot learning} has been introduced to provide a more realistic and practical setting than zero-shot learning, as models are evaluated on both seen and unseen classes~\cite{chao2016empirical}. This change in evaluation has a large impact on existing compatibility functions designed for zero-shot learning, as they do not perform well in the generalized setting~\cite{chao2016empirical,xian2018zero,changpinyo2020classifier}. Indeed, whether they are based on a ranking loss~\cite{akata2016label,akata2015evaluation,xian2016latent,romera2015embarrassingly,frome2013devise} or synthesis~\cite{changpinyo2020classifier,changpinyo2016synthesized,changpinyo2017predicting}, compatibility functions empirically exhibit a very low accuracy for unseen classes. As identified by Chao~\etal~\cite{chao2016empirical}, this indicates a strong inherent bias in all classifiers towards the seen classes.
To overcome the low accuracy for unseen classes, both Kumar Verma~\etal~\cite{kumar2018generalized} and Xian~\etal~\cite{xian2018feature} learn a conditional generative model to generate image features. Once trained, image features of unseen classes are sampled by changing the conditioning of the generative models. Classification then consists of training a one-hot softmax classifier on both real and sampled image features. Having access during training to generated unseen features leads to an increase in unseen class accuracy.
Among the different generative models used in generalized zero-shot learning are generative adversarial networks~\cite{xian2018feature,li2019leveraging,felix2018multi}, variational autoencoders~\cite{schonfeld2018generalized,kumar2018generalized} or a combination of both~\cite{xian2019f}. Still, a classifier trained on generated features suffers from a bias towards seen classes because generative models do not fully match the true distribution of unseen classes. In this paper, we strive for a bias-aware classifier, which can behave as a stand-alone model like compatibility functions and also leverage unseen features sampled from a generative model.

\paragraph{Addressing the bias} in classifiers remains an open challenge for generalized zero-shot learning. Although Chao~\etal~\cite{chao2016empirical} identify the critical bias towards seen classes, only a few works try to address it during training.
Related works separate the seen and unseen classifications.
Liu~\etal~\cite{liu2018generalized} map both features and semantic representations to a common embedding space. Probabilities are then calibrated separately in this common space to make seen class probabilities confident and reduce the uncertainty of unseen class probabilities.
Atzmon and Chechik~\cite{atzmon2018domain} train expert models separately for seen and unseen class predictions. Their predictions are further combined in a soft manner with a third expert to produce the final decision.
In this paper, we strive to address the bias by considering seen and unseen class probabilities jointly rather than separately. Having access during training to the joint class probabilities lets the bias-aware model learn how to balance them from the start.

\section{Method}
\label{sec:metho}

During training, a generalized zero-shot learner $G:X \rightarrow \mathcal{T}$ is given a training set $\mathcal{D}^{\mathcal{S}}=\{(x_n, y_n), y_n\in \mathcal{S}\}_{n=1}^{N}$, where $x_n \in \mathbb{R}^D$ is an image feature of dimension $D$ and $y_n$ comes from the set $\mathcal{S}$ of $\seen$ classes, with $\mathcal{S} \subset \mathcal{T}$.
For each $c \in \mathcal{S}$ there exists a corresponding semantic class representation $\phi(c)\in \mathbb{R}^A$ of dimension $A$. At testing time, $G$ predicts for each sample in the testing set $\mathcal{D}^{\mathcal{T}}=\{x_n\}_{n=1}^{M}$ a label that belongs to $\mathcal{T}$ by exploiting the joint set of $\seen$ and $\unseen$ semantic class representations. This problem formulation can be extended with an auxiliary dataset $\widetilde{\mathcal{D}}^{\mathcal{U}}=\{(\widetilde{x}_n, y_n), y_n\in \mathcal{U}\}_{n=1}^{\widetilde{N}}$, where $y_n$ comes from the set of unseen classes $\mathcal{U}$. $\widetilde{\mathcal{D}}^{\mathcal{U}}$ mimics image features from unseen classes, and is typically sampled from a generative model. The joint set $\{\mathcal{D}^{\mathcal{S}}, \widetilde{\mathcal{D}}^{\mathcal{U}}\}$ now covers both seen and unseen classes.
 
In this paper, we propose a bias-aware generalized zero-shot learner $f(\cdot)$, which can operate during training with only $\mathcal{D}^{\mathcal{S}}$ similar to compatibility functions (Section~\ref{sec:metho:seen}) or the joint set $\{\mathcal{D}^{\mathcal{S}}, \widetilde{\mathcal{D}}^{\mathcal{U}}\}$ similar to classifiers in the generative approach (Section~\ref{sec:metho:unseen}). In both scenarios, the learner includes mechanisms to mitigate the bias towards seen classes.
Learning consists of mapping inputs $x$ to their corresponding semantic class representations $\phi(c)$. In other words, the model regresses to a real-valued vector, which describes a class prototype. 
We denote the set of seen class prototypes as $\Phi^\mathcal{S}=\{\phi(c), c \in \mathcal{S}\}$, unseen class prototypes as $\Phi^\mathcal{U}=\{\phi(c), c \in \mathcal{U}\}$, and their union as $\Phi^\mathcal{T}=\Phi^\mathcal{S}\cup\Phi^\mathcal{U}=\{\phi(c), c \in \mathcal{T}\}$. Usually, the semantic knowledge used for class prototypes corresponds to semantic attributes~\cite{farhadi2009describing,lampert2014attribute}, word vectors of the class name~\cite{palatucci2009zero,frome2013devise}, hierarchical representations~\cite{akata2016label,akata2015evaluation,xian2016latent}, or sentence descriptions~\cite{reed2016learning,xian2018feature}.
To exploit this diversity in semantic knowledge, we propose to swap the representation types for seen and unseen prototypes (Section~\ref{sec:metho:swap}). 

\subsection{Stand-alone classification with seen classes only}\label{sec:metho:seen}
We design the bias-aware generalized zero-shot learner as a probabilistic model with two key principles. First, it is calibrated towards $\seen$ classes such that inputs from $\unseen$ classes yield a low confidence prediction at testing time. In return, this reduces the bias towards seen classes for unseen class inputs. Second, it maps inputs to class prototypes in the semantic embedding space. Following these two principles, we propose:
\begin{equation}
\label{eq:pseen}
    p(c|x,\mathcal{S}) = \exp\left(\dfrac{s\left(f(x), \phi(c)\right)}{T}\right) \bigg/ \sum_{c' \in \mathcal{S}}\exp\left(\dfrac{s\left(f(x), \phi(c')\right)}{T}\right),
\end{equation}
where $s(\cdot, \cdot)$ is the cosine similarity and $T \in \mathbb{R}_{>0}$ is the temperature scale.
When $T = 1$, it acts as the normal softmax function. When $T > 1$,  probabilities are spreading out. When $T < 1$, probabilities tend to concentrate similar to a Dirac delta function. Contrary to knowledge distillation~\cite{hinton2014distilling}, we seek to concentrate the probabilities with a low temperature scale for discriminative purposes.
Learning the probabilistic model is done via minimizing the cross-entropy loss function over the training set of seen examples $\mathcal{D}^{\mathcal{S}}$:
\begin{equation}
\label{eq:loss-seen}
  \mathcal{L}_{\mathrm{s}} = - \frac{1}{N}\sum_{n=1}^{N} \log p(y_n|x_n,\mathcal{S}).
\end{equation}
This probabilistic model behaves like a compatibility function, because it only sees samples from $\seen$ classes during training. At testing, the evaluation simply measures the similarity in the embedding space with respect to the union of seen and unseen prototypes $\Phi^\mathcal{T}$.

Variants of this prototype-based learner have been proposed in image retrieval~\cite{liu2017sphereface,movshovitz2017no,wen2016discriminative,zhai2018making} or image classification~\cite{liu2018generalized,wu2018improving,snell2017prototypical}. We differ by (\textit{i}) fixing the prototypes to be semantic class representations rather than learning them; (\textit{ii}) learning a mapping from the inputs to the class representations rather than learning a common embedding space; (\textit{iii}) applying a softmax function to provide a probabilistic interpretation of cosine similarities; and (\textit{iv}) calibrating the model with the same temperature scaling for both training and testing. 

\subsection{Classification with both seen and unseen classes}\label{sec:metho:unseen}

In the generative approach for generalized zero-shot learning, samples from unseen classes are generated. We can then use the generated data $\widetilde{\mathcal{D}}^{\mathcal{U}}$ as an auxiliary dataset for calibration and for entropy regularization.
In this context, given an input $x$ the probabilistic model learns to predict a class from the union of both $\seen$ and $\unseen$ classes:
\begin{equation}
\label{eq:punseen}
  p(c|x, \mathcal{T}) = \exp\left(\dfrac{s\left(f(x), \phi(c)\right)}{T}\right) \bigg/ \sum_{c' \in \mathcal{T}}\exp\left(\dfrac{s\left(f(x), \phi(c')\right)}{T}\right).
\end{equation}
The only and major difference with eq.~\ref{eq:pseen} resides in the class prototypes that are considered to produce the prediction, while $f(\cdot)$ remains the same model. $p(c|x,\mathcal{S})$ only evaluates over the set of $\seen$ class prototypes $\Phi^\mathcal{S}$, while $p(c|x,\mathcal{T})$ evaluates over the union of seen and unseen class prototypes $\Phi^\mathcal{T}$. In this case, the temperature scaling ensures the model is confident for both seen and unseen classes. This difference also makes the learning distinctive from related works (\textit{i.e.},~DCN~\cite{liu2018generalized} or COSMO~\cite{atzmon2018domain}), as they consider seen and unseen classifications separately rather than jointly.
Akin to eq.~\ref{eq:loss-seen}, we minimize the cross-entropy loss function on the joint set $\{\mathcal{D}^{\mathcal{S}}, \widetilde{\mathcal{D}}^{\mathcal{U}}\}$ of $\seen$ and $\unseen$ classes:
\begin{equation}
\label{eq:loss-unseen}
  \mathcal{L}_{\mathrm{s}+\mathrm{u}} =  - \frac{1}{N}\sum_{n=1}^{N} \log p(y_n|x_n,\mathcal{T}) - \frac{1}{\widetilde{N}}\sum_{n=1}^{\widetilde{N}} \log p(y_n|\widetilde{x}_n,\mathcal{T}).
\end{equation}
This probabilistic model behaves like a classifier used in generative approaches, because it sees samples from both $\seen$ and $\unseen$ classes at both training and testing times, and the partition function normalizes over the union of seen and unseen sets of classes. Having a classification over the union enables regularization in both seen and unseen directions.

\paragraph{Bidirectional entropy regularization.}

Intuitively, when an image from an unseen class is fed to the classifier, probabilities for seen classes should yield a high entropy, while probabilities for unseen classes should result in a low entropy. In other words, the evaluation over seen classes of an unseen class input should be uncertain, because the image comes from a class the classifier has never encountered during training. Conversely, when an image from a seen class is fed to the classifier, the entropy of the probabilities for unseen classes should be high, while the entropy for seen classes should be low. To encourage this effect, given an image $x$, we compute the normalized Shannon entropy~\cite{shannon1948mathematical} of the probabilistic model $p(c|x,\mathcal{T})$ for both seen and unseen class directions:
\begin{align}
  \mathcal{H}_{\mathrm{s}}(x) = \frac{-1}{|\mathcal{S}|} \sum_{c \in \mathcal{S}} p(c|x,\mathcal{T}) \log p(c|x,\mathcal{T})\textrm{, and~}
  \mathcal{H}_{\mathrm{u}}(x) = \frac{-1}{|\mathcal{U}|} \sum_{c \in \mathcal{U}} p(c|x,\mathcal{T}) \log p(c|x,\mathcal{T}),
\end{align}
where $\mathcal{H}_{\mathrm{s}}$ and $\mathcal{H}_{\mathrm{u}}$ are the average entropy for seen and unseen classes, and $|\cdot|$ is the cardinality of the set. For training, we derive a margin-based regularization for both seen and unseen class directions:
\begin{align}
  R_{\mathrm{s}} = \left[ m + \frac{1}{N}\sum_{n=1}^{N} \mathcal{H}_{\mathrm{s}}(x_n) - \frac{1}{\widetilde{N}}\sum_{n=1}^{\widetilde{N}} \mathcal{H}_{\mathrm{s}}(\widetilde{x}_n) \right]_+,\\
  R_{\mathrm{u}} = \left[ m + \frac{1}{\widetilde{N}}\sum_{n=1}^{\widetilde{N}} \mathcal{H}_{\mathrm{u}}(\widetilde{x}_n) - \frac{1}{N}\sum_{n=1}^{N} \mathcal{H}_{\mathrm{u}}(x_n) \right]_+,
\end{align}
where $[\cdot ]_+ = \max(0, \cdot)$. $R_{\mathrm{s}}$ ensures a margin of at least $m$ between the average seen class entropy of seen inputs $x_n$ and generated unseen inputs $\widetilde{x}_n$. In other words, this formulation seeks to minimize $\mathcal{H}_{\mathrm{s}}(x_n)$ and maximize $\mathcal{H}_{\mathrm{s}}(\widetilde{x}_n)$. $R_{\mathrm{u}}$ has a corresponding effect on the unseen class entropy.
The final loss function for training then becomes:
\begin{equation}
\label{eq:loss-su}
  \mathcal{L}_{\mathrm{f}} = \mathcal{L}_{\mathrm{s+u}} + \lambda_{\mathrm{Ent}} (R_\mathrm{s} + R_\mathrm{u}),
\end{equation}
where $\lambda_{\mathrm{Ent}} \in \mathbb{R}_{\geq0}$ is a hyper-parameter to control the contribution of the bidirectional entropy.

\subsection{Swapping seen and unseen class representations}\label{sec:metho:swap}

As presented above, relying on a real-valued embedding space allows mechanisms to mitigate the bias in two scenarios. It also enables to swap class representations to less biased representations.
Consider now the case where there exist multiple types of semantic information, which differ by their type of representation and by how expensive it is to collect them. For example, attribute descriptions require expert knowledge, while sentence descriptions can be crowd-sourced to non-expert workers. Practically, sentences tend to be less biased than attributes and perform better~\cite{xian2018feature}, but do not offer a comprehensive expert-based explanation~\cite{reed2016learning}.
One could then train a model for seen classes on attributes as they rely on expert-based explanations and rely for unseen classes on sentences as they are easier to collect. This results in different representation types for seen and unseen classes.

Formally, we assume that we have access to seen prototypes $\{\Phi_A^\mathcal{S}, \Phi_B^\mathcal{S}\}$ with representations from domain $A$ and $B$. For evaluation, we have access to unseen prototypes $\Phi_A^\mathcal{U}$ of domain $A$, but $\Phi_B^\mathcal{U}$ of domain $B$ is absent.
The objective is then to learn a mapping $\beta$ from $\Phi_A^\mathcal{S}$ to $\Phi_B^\mathcal{S}$, in order to regress $\hat{\Phi}_B^\mathcal{U}$ from $\Phi_A^\mathcal{U}$ at testing time.
We define the mapping as a linear least squares regression problem with Tikhonov regularization, which corresponds to:
\begin{equation}
\label{eq:lsq}
  \min_{\beta}\lVert\Phi_B^\mathcal{S} - \beta\Phi_A^\mathcal{S}\rVert_2 + \lambda_\beta \lVert \beta\rVert_2.
\end{equation}
where $\lambda_\beta$ controls the amount of regularization. Relying on a linear transformation prevents overfitting, as the mapping involves a limited set of class prototypes.
During evaluation, we apply $\beta$ to unseen prototypes of domain $A$ to regress their values in domain $B$: $\hat{\Phi}_B^\mathcal{U} = \beta\Phi_A^\mathcal{U}$. Swapping representations then corresponds to regressing from one domain to another.

\section{Experimental Details}
\label{sec:details}

\paragraph{Datasets.} We report experiments on four datasets commonly used in generalized zero-shot learning, 
\eg,~\cite{chao2016empirical,changpinyo2020classifier,xian2018zero,reed2016learning}. For all datasets, we rely on the train and test splits proposed by Xian~\etal~\cite{xian2018zero}.
\textit{\textbf{Caltech-UCSD-Birds 200-2011 (CUB)}}~\cite{WahCUB_200_2011} contains 11,788 images from 200 bird species. Every species is described by a unique combination of 312 semantic attributes to characterize the color, pattern and shape of their specific parts. Moreover, every bird image comes along with 10 sentences describing the most prominent characteristics~\cite{reed2016learning}. 150 species are used as $\seen$ classes during training, and 50 distinct species are left out as $\unseen$ classes during testing.
\textit{\textbf{SUN Attribute (SUN)}}~\cite{Patterson2012SunAttributes} contains 14,340 images from 717 scene types. Every scene is also described by a unique combination of 102 semantic attributes to characterize material and surface properties. 645 scene types are used as $\seen$ classes during training, and 72 distinct scene types are left out as $\unseen$ classes during testing.
\textit{\textbf{Animals with Attributes (AWA)}}~\cite{lampert2014attribute} contains 30,475 images from 50 animals. Every animal comes with a unique combination of 85 semantic attributes to describe their color, shape, state or function. 40 animals are used as $\seen$ classes during training, and 10 distinct animals are left out as $\unseen$ classes during testing.
\textit{\textbf{Oxford Flowers (FLO)}}~\cite{nilsback2008automated} contains 8,189 images from 102 flower plants. Every flower plant image is described by 10 different sentences describing the shape and appearance~\cite{reed2016learning}. 82 flowers are used as $\seen$ classes during training, and 20 distinct flowers are left out as $\unseen$ classes during testing.

\paragraph{Features extraction.}
For all datasets, we rely on the features extracted by Xian~\etal~\cite{xian2018zero}. Image features $x$ come from ResNet101~\cite{he2016deep} trained on ImageNet~\cite{ILSVRC15} and sentence representations are extracted from a 1024-dimensional CNN-RNN~\cite{reed2016learning}. As established by Xian~\etal~\cite{xian2018zero}, parameters of ResNet101 and the CNN-RNN are frozen and are not fine-tuned during the training phase. No data augmentation is performed either.

\paragraph{Evaluation.} We evaluate experiments with calibration stacking as proposed by Chao~\etal~\cite{chao2016empirical}, which penalizes the seen class probabilities to reduce the bias during evaluation. Following Xian~\etal~\cite{xian2018zero}, we compute the average per-class top-1 accuracy of $\seen$ classes (denoted as $\mathbf{s}$) and $\unseen$ classes (denoted as $\mathbf{u}$), as well as their harmonic mean $\textbf{H} = (2\times \mathbf{s} \times \mathbf{u}) / (\mathbf{s} + \mathbf{u})$. We report the 3-run average.

\paragraph{Implementation details.}
In our model, $f(\cdot)$ corresponds to a multilayer perceptron with 2 hidden layers of size 2048 and 1024 to map the features $x$ to the joint visual-semantic embedding space of size $A$. The output layer has a linear activation, while hidden layers have a ReLU activation~\cite{nair2010rectified} followed by a Dropout regularization ($p=0.5$)~\cite{srivastava2014dropout}.
We train $f(\cdot)$ using stochastic gradient descent with Nesterov momentum~\cite{sutskever2013importance}.
We set the following hyper-parameters for all datasets: learning rate of 0.01 with cosine annealing~\cite{loshchilov2017sgdr}, initial momentum of 0.9, batch size of 64, temperature of 0.05, and an entropy regularization term of 0.1 with a margin of 0.2.
For AWA, we reduce the learning rate to 0.0001 and increase the entropy regularization to 0.5 while keeping the same margin.
When relying on sentence representations, we double the capacity of $f(\cdot)$ with twice the number of hidden units in each layer.
We set hyper-parameters on a hold-out validation set and re-train on the joint training and validation sets.
The source code uses the Pytorch framework~\cite{paszke2019pytorch}.

\begin{figure}[t]
\centering
\begin{minipage}{.48\textwidth}
\centering
\includegraphics[width=.9\linewidth]{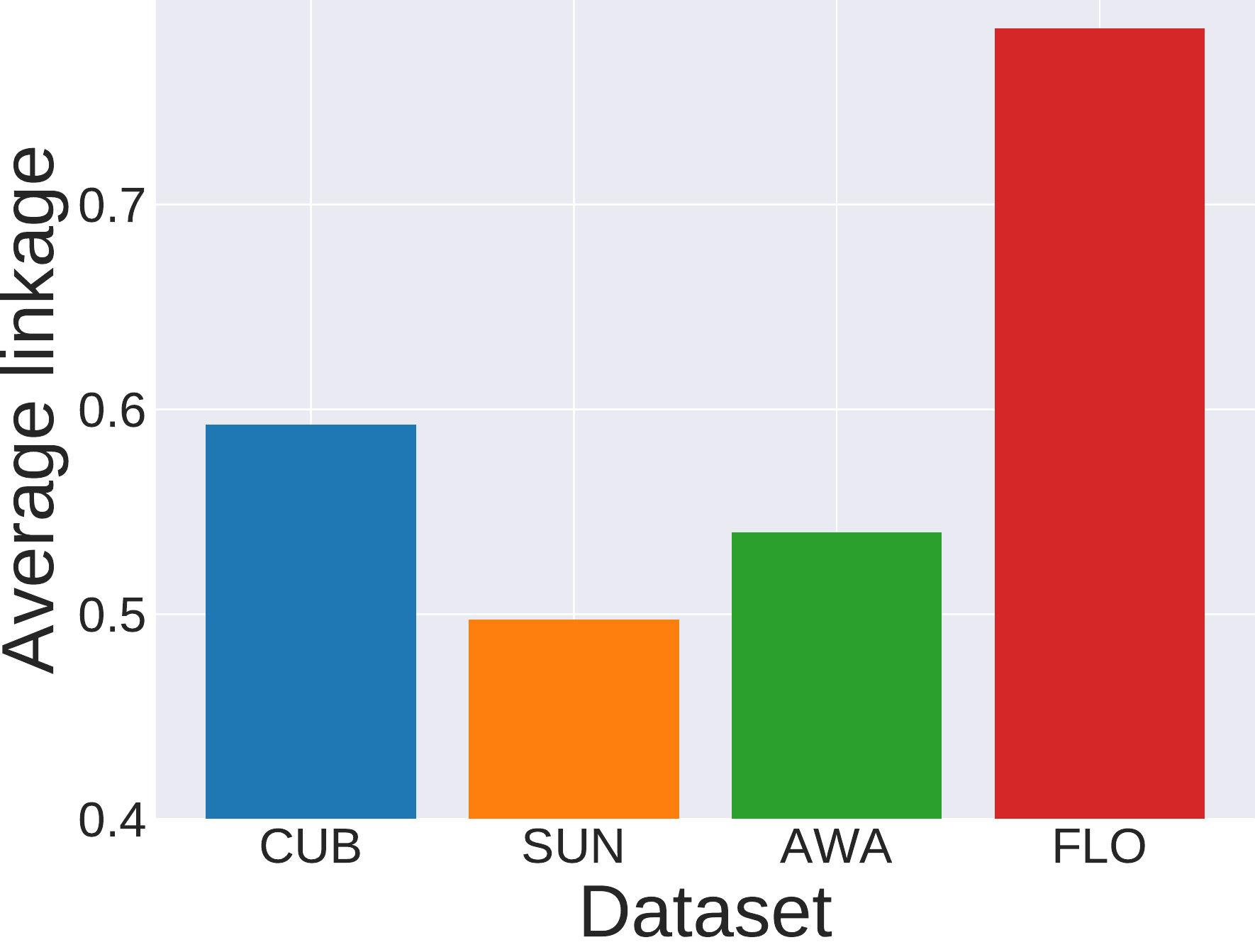}
\caption{\textbf{Bias variation} across datasets. When measuring the average linkage between seen and unseen representations, FLO is the most affected while SUN is the least. Thus, the bias towards seen classes differs across datasets.}
\label{fig:linkage}
\end{minipage}\hfill
\begin{minipage}{.48\textwidth}
\centering
\vspace{7pt}
\hfill
\begin{overpic}[width=0.30\linewidth]{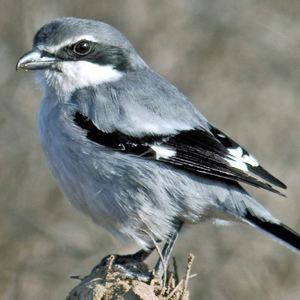}
\put(-15,30){\rotatebox{90}{Seen}}
\put(35,103){\small CUB}
\put(22,-9){\footnotesize \textit{great grey}}
\end{overpic}~
\begin{overpic}[width=0.30\linewidth]{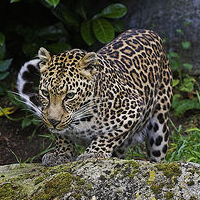}
\put(30,103){\small AWA}
\put(25,-9){\footnotesize \textit{leopard}}
\end{overpic}~
\begin{overpic}[width=0.30\linewidth]{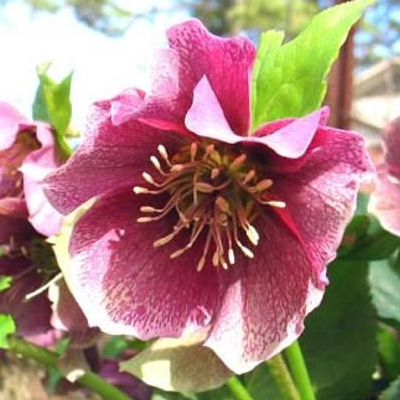}
\put(30,103){\small FLO}
\put(20,-9){\footnotesize \textit{leten rose}}
\end{overpic}\\[6pt]

\hfill
\begin{overpic}[width=0.30\linewidth]{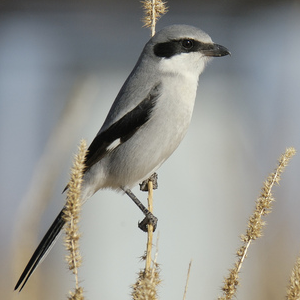}
\put(-15,21){\rotatebox{90}{Unseen}}
\put(21,-9){\footnotesize \textit{loggerhead}}
\end{overpic}~
\begin{overpic}[width=0.30\linewidth]{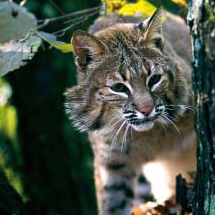}
\put(28,-9){\footnotesize \textit{bobcat}}
\end{overpic}~
\begin{overpic}[width=0.30\linewidth]{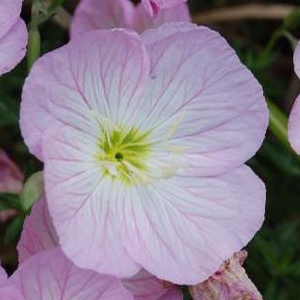}
\put(8,-9){\footnotesize \textit{pink primrose}}
\end{overpic}\\[1pt]

\caption{\textbf{Seen and unseen class samples}. Visual differences arise from the global shape (CUB, AWA) or colors (FLO). Yet, their semantic class representation yields a very high pairwise similarity, which creates a high bias.\newline}
\label{fig:ex}
\end{minipage}
\end{figure}

\section{Results}
\label{sec:res}

\paragraph{Bias variation.} To verify whether the bias towards seen classes is dataset-dependent, we measure the average linkage between seen and unseen representations. Concretely, we compute the average of the pairwise cosine similarity between $\Phi^{\mathcal{S}}$ and $\Phi^{\mathcal{U}}$.
A high average linkage then refers to a high similarity between seen and unseen representations. Intuitively, a high average linkage is not desirable as unseen representations can easily be confused with seen ones, which makes the generalized zero-shot learning problem harder. 
Figure~\ref{fig:linkage} depicts the average linkage per dataset. FLO exhibits the highest average linkage while SUN the lowest, with a 1.6 times difference. In other words, classifiers trained on FLO are highly affected by the bias towards seen classes.
Figure~\ref{fig:ex} illustrates seen and unseen class samples with a very high pairwise similarity on CUB, AWA and FLO. Visually, these classes can be differentiated by their color or shape. Though, their semantic representations are very similar, which creates a high bias.
Now that we have established that the bias towards seen classes differs across datasets, we can address the bias within generalized zero-shot learners.

\begin{figure}[t]
\centering
\begin{minipage}{.48\textwidth}
\centering
\includegraphics[width=.9\linewidth]{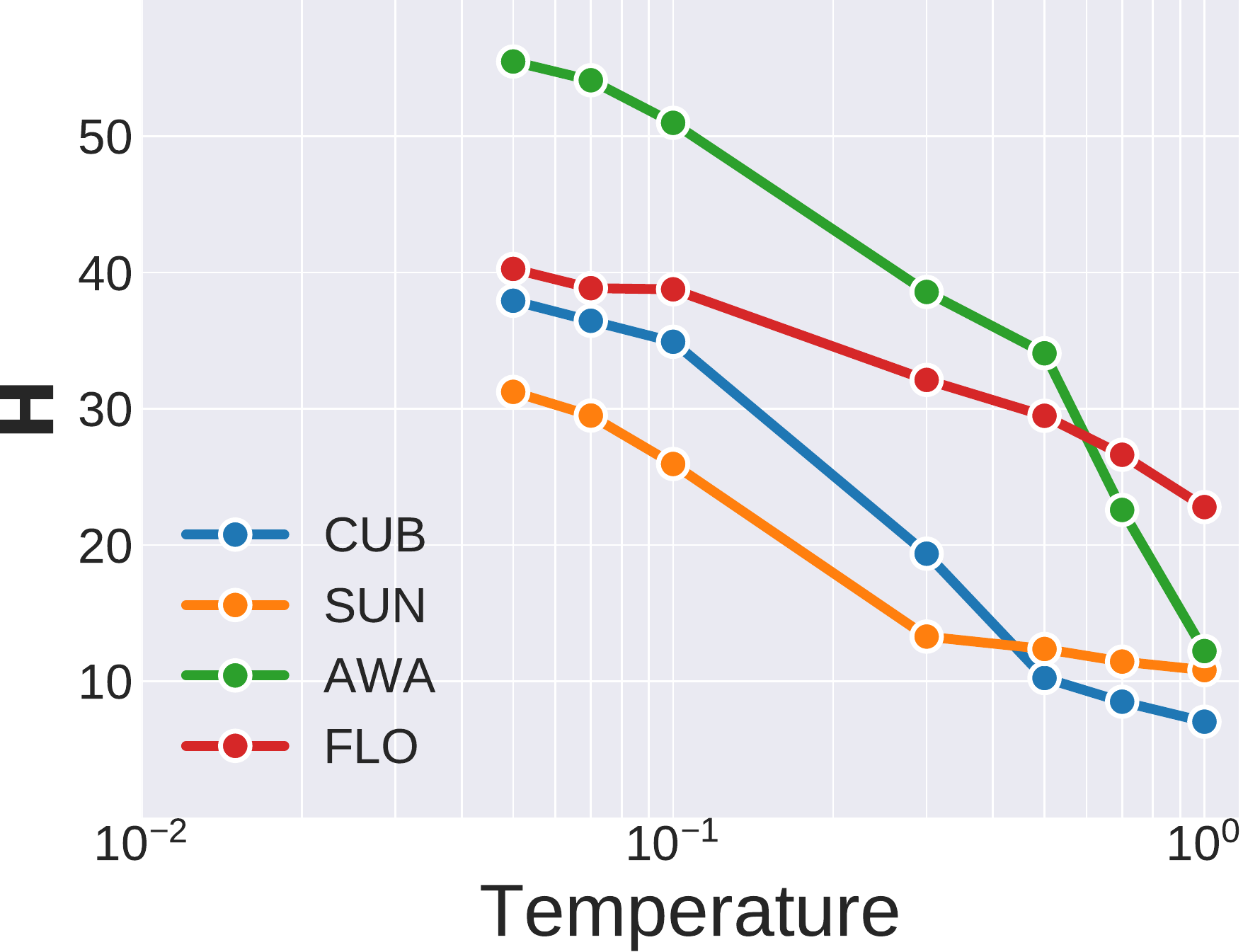}
\caption{\textbf{Temperature scaling} ablation from $T=0.05$ to $T=1$. Temperature values over 0.1 degrade the performance because probabilities start to spread which makes the model less confident.\newline\newline}
\label{fig:temp}
\end{minipage}\hfill
\begin{minipage}{.48\textwidth}
\centering
\includegraphics[width=.94\linewidth]{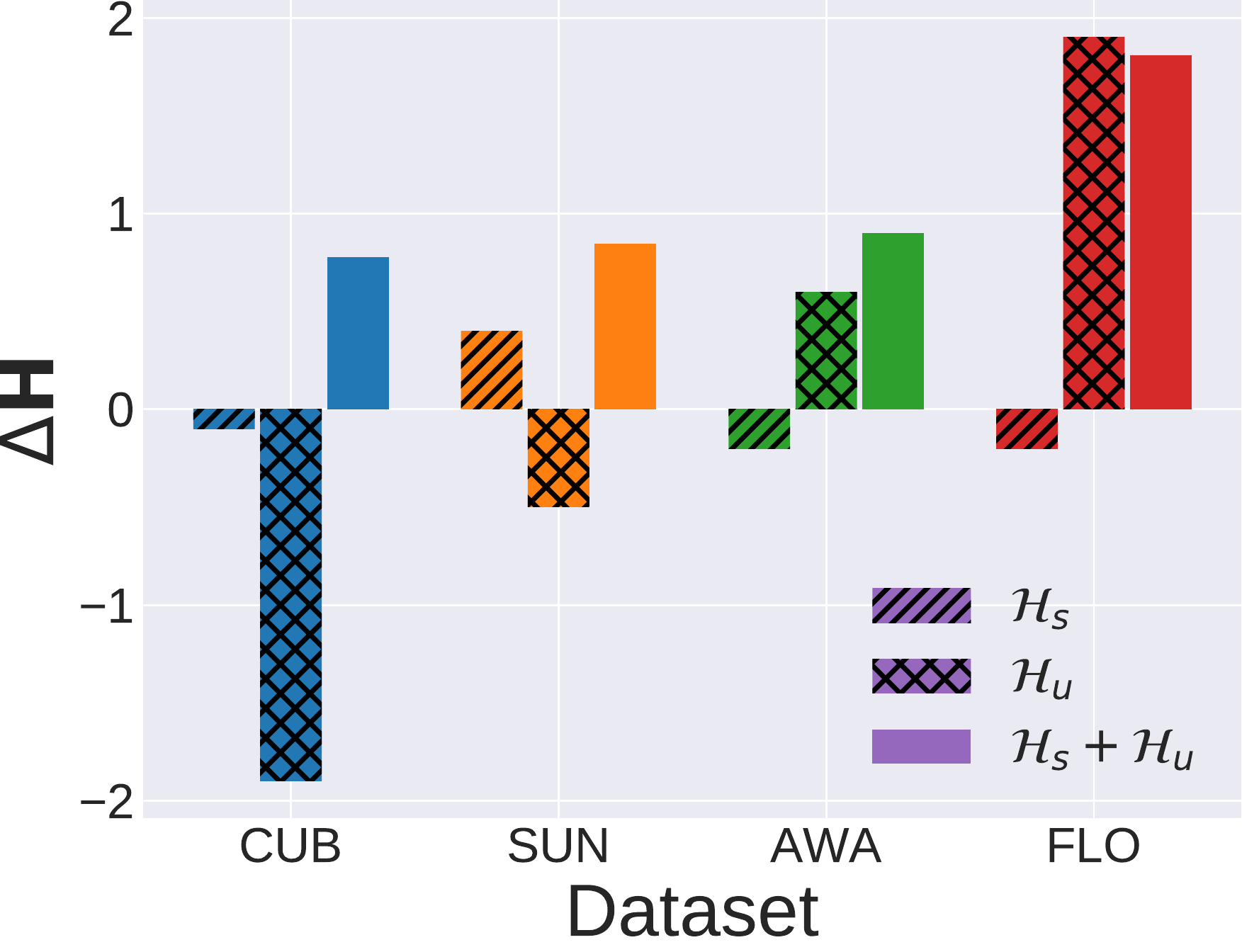}
\caption{\textbf{Entropy regularization} in one ($\mathcal{H}_s$ or $\mathcal{H}_u$, hatched) and two ($\mathcal{H}_s + \mathcal{H}_u$, not hatched) directions compared with models without. Regularizing in only one direction can result in a negatively effect. Including both directions consistently improves results by creating a better bias trade-off.}
\label{fig:bias}
\end{minipage}
\end{figure}

\paragraph{Temperature scaling.} Figure~\ref{fig:temp} varies the scale of the temperature in eq.~\ref{eq:pseen}. Following related metric learning works (\eg,~\cite{wu2018improving,zhai2018making}), we consider the temperature as a hyper-parameter. When treated as a latent parameter, the optimization diverges as its value goes down to zero to satisfy the loss function. The highest \textbf{H} score occurs when $T=0.05$ on the validation set of all datasets. Performance starts to degrade substantially after $T>0.1$. A temperature lower than $T<0.05$ can yield even higher scores, but is usually prone to numerical errors. As such, we set $T=0.05$ in all our experiments when training the model with only seen samples (eq.~\ref{eq:loss-seen}) or in combination with generated unseen samples (eq.~\ref{eq:loss-unseen}).
We also evaluate modifying $T$ between training and testing phases. Setting it to 1 during training and testing, as in a normal softmax, drops \textbf{H} by 43.3\% on AWA. And changing it to 0.05 when testing, drops the score by 25.6\%.
Keeping a fixed temperature value ensures $f(\cdot)$ maps inputs to prototypes similarly in training and testing. The temperature value should also be low to promote a more confident and discriminative model that yields narrow probabilities. Hence, the model reduces the bias by having a lower likelihood to classify an unseen class input as part of a seen class.

\paragraph{Entropy regularization.} Figure~\ref{fig:bias} ablates the direction of the margin-based entropy term in eq.~\ref{eq:loss-su}. For this experiment, we rely on unseen class features generated from Cycle-CLSWGAN~\cite{felix2018multi}. When using a unidirectional entropy regularization, the improvement is either very low, or even negative, over a model without any regularization. Interestingly, this negative effect does not depend on the direction, as both $\mathcal{H}_s$ and $\mathcal{H}_u$ are affected when considered individually. Regularizing in only one direction forces the model to compensate for the other direction.
Only the bidirectional regularization provides a consistent benefit for all datasets. This positive effect indicates the importance of balancing out both seen and unseen probabilities when mitigating the bias. Regularizing in both directions jointly helps the model learn a correct bias trade-off.

\paragraph{Swapping representations.} Table~\ref{tab:swap} presents the different combinations of attribute (\texttt{Att}) and sentence (\texttt{Sen}) representations for training and evaluation. \texttt{Att}-\texttt{Att} and \texttt{Sen}-\texttt{Sen} are the common non-swapped settings. \texttt{Sen}-\texttt{Sen} forms an upper-bound as sentences provide better class representations over attributes. Indeed, sentence descriptions exhibit a lower average linkage than attribute descriptions.
In a swapped setting, the unseen representations are regressed from representations in another domain based on eq.~\ref{eq:lsq}.
A model trained on \texttt{Att} can be improved by 1.2 points at testing time when using \texttt{Sen} to regress the unseen representations. However, a model trained on \texttt{Sen} degrades when using \texttt{Att} to regress unseen representations. Indeed, \texttt{Sen}-\texttt{Att} requires to map low-dimensional attribute representations of unseen classes to a high-dimensional space of sentence representations on which the classifier has been trained. \texttt{Sen}-\texttt{Att} then involves dimensionality expansion, which is a harder problem than dimensionality compression in \texttt{Att}-\texttt{Sen}.
In the scenario where a model is trained on attributes for seen classes derived from experts, it is possible to leverage sentences for unseen classes derived from crowd-sourcing to further improve the results.

\begin{table}[t]
\centering
\begin{minipage}{.3\textwidth}
\centering
\tablestyle{6pt}{1.}
  \begin{tabular}{ccc}
  \toprule
  Seen & Unseen & \textbf{H}\\
  \cmidrule(lr){1-2}\cmidrule(lr){3-3}
  \texttt{Att} & \texttt{Att} & 48.5 \\
  \texttt{Sen} & \texttt{Att} & 47.4 \\
  \texttt{Att} & \texttt{Sen} & 49.7 \\
  \texttt{Sen} & \texttt{Sen} & 50.3 \\
  \bottomrule
  \end{tabular}  

\end{minipage}
\begin{minipage}{.68\textwidth}
\caption{\textbf{Swapping attribute (\texttt{Att}) and sentence (\texttt{Sen}) representations}. While \texttt{Att}-\texttt{Att} and \texttt{Sen}-\texttt{Sen} are the usual non-swapped evaluation settings, our method can also swap them. When using sentences for unseen classes, it always improves upon attributes in swapped and non-swapped evaluations as they are less biased and more discriminative.}
\label{tab:swap}
\end{minipage}
\end{table}

\begin{table}[h!]
  \centering
\tablestyle{2.7pt}{1.0}
  \begin{tabular}{lcccccccccccc}
    \toprule
    \multirow{2}{*}{Method} & \multicolumn{3}{c}{\textbf{CUB}} & \multicolumn{3}{c}{\textbf{SUN}} & \multicolumn{3}{c}{\textbf{AWA}} & \multicolumn{3}{c}{\textbf{FLO}} \\
    \cmidrule(lr){2-13}
    & $\mathbf{u}$ & $\mathbf{s}$ & \cellcolor{Gray}$\mathbf{H}$ & $\mathbf{u}$ & $\mathbf{s}$ & \cellcolor{Gray}$\mathbf{H}$ & $\mathbf{u}$ & $\mathbf{s}$ & \cellcolor{Gray}$\mathbf{H}$ & $\mathbf{u}$ & $\mathbf{s}$ & \cellcolor{Gray}$\mathbf{H}$\\

    \midrule
    DeViSE~\cite{frome2013devise} & 23.8 & 53.0 & \cellcolor{Gray}32.8 & 16.9 & 27.4 & \cellcolor{Gray}20.9 & 13.4 & 68.7 & \cellcolor{Gray}22.4 & 9.9 & 44.2 & \cellcolor{Gray}16.2 \\
    \texttt{w/ f-CLSWGAN}~\cite{xian2018feature} & 52.2 & 42.4 & \cellcolor{Gray}46.7 & 38.4 & 25.4 & \cellcolor{Gray}30.6 & 35.0 & 62.8 & \cellcolor{Gray}45.0 & 45.0 & 38.6 & \cellcolor{Gray}41.6\\
    
    \midrule
    SJE~\cite{akata2015evaluation} & 23.5 & 59.2 & \cellcolor{Gray}33.6 & 14.7 & 30.5 & \cellcolor{Gray}19.8 & 11.3 & 74.6 & \cellcolor{Gray}19.6 & 13.9 & 47.6 & \cellcolor{Gray}21.5 \\
    \texttt{w/ f-CLSWGAN}~\cite{xian2018feature} & 48.1 & 37.4 & \cellcolor{Gray}42.1 & 36.7 & 25.0 & \cellcolor{Gray}29.7 & 37.9 & 70.1 & \cellcolor{Gray}49.2 & 52.1 & 56.2 & \cellcolor{Gray}54.1\\
    
    \midrule
    LATEM~\cite{xian2016latent} & 15.2 & 57.3 & \cellcolor{Gray}24.0 & 14.7 & 28.8 & \cellcolor{Gray}19.5 & 7.3 & 71.7 & \cellcolor{Gray}13.3 & 6.6 & 47.6 & \cellcolor{Gray}11.5\\
    \texttt{w/ f-CLSWGAN}~\cite{xian2018feature} & 53.6 & 39.2 & \cellcolor{Gray}45.3 & 42.4 & 23.1 & \cellcolor{Gray}29.9 & 33.0 & 61.5 & \cellcolor{Gray}43.0 & 47.2 & 37.7 & \cellcolor{Gray}41.9 \\

    \midrule
    ESZSL~\cite{romera2015embarrassingly} & 12.6 & 63.8 & \cellcolor{Gray}21.0 & 11.0 & 27.9 & \cellcolor{Gray}15.8 & 6.6 & 75.6 & \cellcolor{Gray}12.1 & 11.4 & 56.8 & \cellcolor{Gray}19.0 \\
    \texttt{w/ f-CLSWGAN}~\cite{xian2018feature} & 36.8 & 50.9 & \cellcolor{Gray}43.2 & 27.8 & 20.4 & \cellcolor{Gray}23.5 & 31.1 & 72.8 & \cellcolor{Gray}43.6 & 25.3 & 69.2 & \cellcolor{Gray}37.1 \\
    
    \midrule
    ALE~\cite{akata2016label} & 23.7 & 62.8 & \cellcolor{Gray}34.4 & 21.8 & 33.1 & \cellcolor{Gray}26.3 & 16.8 & 76.1 & \cellcolor{Gray}27.5 & 13.3 & 61.6 & \cellcolor{Gray}21.9 \\
    \texttt{w/ f-CLSWGAN}~\cite{xian2018feature} & 40.2 & 59.3 & \cellcolor{Gray}47.9 & 41.3 & 31.1 & \cellcolor{Gray}35.5 & 47.6 & 57.2 & \cellcolor{Gray}52.0  & 54.3 & 60.3 & \cellcolor{Gray}57.1\\
    
    \midrule
    DCN~\cite{liu2018generalized} & 28.4 & 60.7 & \cellcolor{Gray}38.7 & 25.5 & 37.0 & \cellcolor{Gray}30.2 & 25.5 & 84.2 & \cellcolor{Gray}39.1 & \deemph{--} & \deemph{--} & \cellcolor{Gray}\deemph{--}\\
    
    \midrule
    One-hot softmax~\cite{xian2018feature} & \deemph{n/a} & \deemph{n/a} & \cellcolor{Gray}\deemph{n/a} & \deemph{n/a} & \deemph{n/a} & \cellcolor{Gray}\deemph{n/a} & \deemph{n/a} & \deemph{n/a} & \cellcolor{Gray}\deemph{n/a} & \deemph{n/a} & \deemph{n/a} & \cellcolor{Gray}\deemph{n/a}\\
    \texttt{w/ f-CLSWGAN}~\cite{xian2018feature} & 43.7 & 57.7 & \cellcolor{Gray}49.7 & 42.6 & 36.6 & \cellcolor{Gray}39.4 & 57.9 & 61.4 & \cellcolor{Gray}59.6 & 59.0 & 73.8 & \cellcolor{Gray}65.6 \\
    \texttt{w/ Cycle-CLSWGAN}~\cite{felix2018multi}$\dagger$ & 45.7 & 61.0 & \cellcolor{Gray}52.3 & 49.4 & 33.6 & \cellcolor{Gray}40.0 & 56.9 & 64.0 & \cellcolor{Gray}60.2 & 72.5 & 59.2 & \cellcolor{Gray}65.1\\
    \texttt{w/ CADA-VAE}~\cite{schonfeld2018generalized} & 51.6 & 53.5 & \cellcolor{Gray}52.4 & 47.2 & 35.7 & \cellcolor{Gray}40.6 & 57.3 & 72.8 & \cellcolor{Gray}64.1 & \deemph{--} & \deemph{--} & \cellcolor{Gray}\deemph{--}\\
    \texttt{w/ f-VAEGAN-D2}~\cite{xian2019f}$\dagger$ & 48.4 & 60.1 & \cellcolor{Gray}53.6 & 45.1 & 38.0 & \cellcolor{Gray}\textbf{41.3} & 57.6 & 70.6 & \cellcolor{Gray}63.5 & 56.8 & 74.9 & \cellcolor{Gray}64.6\\
    \texttt{w/ LisGAN}~\cite{li2019leveraging}  & 46.5 & 57.9 & \cellcolor{Gray}51.6  & 42.9 & 37.8 & \cellcolor{Gray}40.2 & 52.6  & 76.3 & \cellcolor{Gray}62.3 & 57.7 & 83.9 & \cellcolor{Gray}68.3\\
    
    \midrule
    COSMO~\cite{atzmon2018domain} & \deemph{n/a} & \deemph{n/a} & \cellcolor{Gray}\deemph{n/a} & \deemph{n/a} & \deemph{n/a} & \cellcolor{Gray}\deemph{n/a} & \deemph{n/a} & \deemph{n/a} & \cellcolor{Gray}\deemph{n/a} & \deemph{n/a} & \deemph{n/a} & \cellcolor{Gray}\deemph{n/a} \\
    \texttt{w/ f-CLSWGAN}~\cite{xian2018feature} & 60.5 & 41.0 & \cellcolor{Gray}48.9 & 35.3 & 40.2 & \cellcolor{Gray}37.6 & 64.8 & 51.7 & \cellcolor{Gray}57.5 & 59.6 & 81.4 & \cellcolor{Gray}68.8\\
    \texttt{w/ LAGO}~\cite{atzmon2018probabilistic} & 44.4 & 57.8 & \cellcolor{Gray}50.2 & 44.9 & 37.7 & \cellcolor{Gray}41.0 & 52.8 & 80.0 & \cellcolor{Gray}63.6 &  \deemph{n/a} & \deemph{n/a} & \cellcolor{Gray}\deemph{n/a}\\
    
    \midrule
    \textit{\textbf{This paper}} & 45.1 & 52.5 & \cellcolor{Gray}\underline{48.5} & 41.0 & 30.1 & \cellcolor{Gray}\underline{34.7} & 55.2 & 70.5 & \cellcolor{Gray}\underline{61.9} & 42.6 & 66.6 & \cellcolor{Gray}\underline{52.0}\\
    \texttt{w/ f-CLSWGAN}~\cite{xian2018feature} & 50.7 & 49.9 & \cellcolor{Gray}50.3 & 41.1 & 31.6 & \cellcolor{Gray}35.7 & 57.7 & 68.4 & \cellcolor{Gray}62.5 & 53.8 & 76.0 & \cellcolor{Gray}63.0 \\
    \texttt{w/ Cycle-CLSWGAN}~\cite{felix2018multi}$\dagger$ & 57.4 & 58.2 & \cellcolor{Gray}\textbf{57.8} & 44.8 & 32.7 & \cellcolor{Gray}37.8 & 61.3 & 69.2 & \cellcolor{Gray}\textbf{65.0} & 69.3 & 79.9 & \cellcolor{Gray}\textbf{74.2}\\
    
    \bottomrule
    \multicolumn{12}{l}{\footnotesize$\dagger$ Method relies on sentence representations instead of attribute representations for CUB.}
  \end{tabular}
\caption{\textbf{Comparison with the state of the art}, where classifiers are delimited by a horizontal rule and their combination with a generative model is in \texttt{teletype} font. ``n/a'' denotes a non-applicable setting to the method while ``--'' refers to non-reported results in the original paper. 
Compared with one-hot softmax and COSMO, our proposal is a stand-alone method that can also operate with seen class samples only. 
Compared with the other compatibility functions that also operate in this similar stand-alone setting, it achieves the best results (\underline{underlined}).
When extended with generated unseen class samples, we also improve over other classifiers (\textbf{bold}), leading to state-of-the-art results on the three most biased datasets out of four (see Figure~\ref{fig:linkage}\label{tab:sota}).
}
\end{table}

\paragraph{Comparison with the state of the art.}
Table~\ref{tab:sota} compares our bias-aware prototype learner with eight other classifiers. Scores from other classifiers correspond to the performance as reported by the authors in their original paper.
First, we consider stand-alone classifiers, which only observe the seen class inputs during training, \textit{i.e.}, without using any generated features. In this setting, our bias-aware formulation outperforms existing compatiblity functions~\cite{frome2013devise,romera2015embarrassingly,akata2016label,akata2015evaluation,xian2016latent,liu2018generalized} on all datasets. It is also interesting to note that recent formulations with one-hot softmax~\cite{xian2018feature} or COSMO~\cite{atzmon2018domain} cannot operate in this setting. Indeed, they rely on a discrete label space for classification while we rely on a real-valued embedding space. This enables our formulation to incorporate new unseen classes easily and at near zero cost, similar to compatibility functions.
Second, our approach is easily extended with existing generative models to include an auxiliary dataset $\widetilde{\mathcal{D}}^{\mathcal{U}}$ for unseen classes. We select f-CLSWGAN~\cite{xian2018feature} and Cycle-CLSWGAN~\cite{felix2018multi} as the authors provide source code to evaluate on all four datasets. Reproducing the models from their original source code yields results within a reasonable range, \textit{i.e.}, less than a 2-point difference in the \textbf{H} score.
We obtain better results with Cycle-CLSWGAN~\cite{felix2018multi} than f-CLSWGAN~\cite{xian2018feature}, which highlights the importance of the quality of the generated unseen class features. Moreover, our method profits more when generated samples better reflect the true distribution. When switching from f-CLSWGAN~\cite{xian2018feature} to Cycle-CLSWGAN~\cite{felix2018multi} on CUB, a one-hot softmax classifier leads to a 2.6\% increase while our bias-aware classifier with a joint entropy regularization yields a 7.5\% increase.
We achieve state-of-the-art results on CUB, AWA and FLO. Only on the SUN dataset the one-hot softmax~\cite{xian2018feature} and COSMO~\cite{atzmon2018domain} provide higher scores. This originates from a lower bias towards seen classes in the SUN dataset (see Figure~\ref{fig:linkage}), which makes a bias-aware model less beneficial. When a dataset exhibits a low bias, separating the model for seen and unseen classes is preferred for equal treatment. Conversely, when a dataset exhibits a high bias, the training of the model should consider seen and unseen classes jointly to balance out their probabilities from the start.
Overall, we produce competitive results in both scenarios, especially compared with classifiers without any bias-awareness.

\section{Conclusion}
The classification of seen and unseen classes in generalized zero-shot learning requires models to be aware of the bias towards seen classes. In this paper, we present such a model which calibrates the probabilities of seen and unseen classes jointly during training, and ensures a margin between the average entropy of both seen and unseen class probabilities.
Learning consists of regressing inputs to real-valued representations. Relying on a mapping to a real-valued embedding space enables to swap seen and unseen representation types, and to evaluate the model in a stand-alone scenario or in combination with generated unseen features.
Overall, our proposed bias-aware learner provides an effective alternative to separate classification approaches or classifiers without bias-awareness.

\vspace{0.5em}
\paragraph{Acknowledgements} {The authors thank Zeynep Akata for helpful initial discussions, and Hubert Banville for feedback. William Thong is partially supported by an NSERC scholarship.}

\bibliography{egbib}
\end{document}